\documentclass[conference]{IEEEtran}
\IEEEoverridecommandlockouts

\usepackage{siunitx}

\usepackage{cite}
\usepackage{amsmath,amssymb,amsfonts}
\usepackage{algorithmic}
\usepackage{graphicx}
\usepackage{textcomp}
\usepackage{xcolor}
\usepackage{url}
\def\BibTeX{{\rm B\kern-.05em{\sc i\kern-.025em b}\kern-.08em
    T\kern-.1667em\lower.7ex\hbox{E}\kern-.125emX}}

\begin{document}

\title{A Systematic Evaluation of Machine Learning Methods for Fault Detection and Line Identification in  Electrical Power Grids
}

\author{
\IEEEauthorblockN{Julian Oelhaf\textsuperscript{1}\textsuperscript{*}, Georg Kordowich\textsuperscript{2}, Paula Andrea Pérez-Toro\textsuperscript{1}, Tomás Arias-Vergara\textsuperscript{1}, Andreas Maier\textsuperscript{1} \\
Johann Jäger\textsuperscript{2}, Siming Bayer\textsuperscript{1}}
\IEEEauthorblockA{\textit{\textsuperscript{1}Pattern Recognition Lab, Friedrich-Alexander-Universität Erlangen-Nürnberg} \\
\textit{\textsuperscript{2}Institute of Electrical Energy Systems, Friedrich-Alexander-Universität Erlangen-Nürnberg} \\
Erlangen, Germany \\
\textsuperscript{*}corresponding author: julian.oelhaf@fau.de}
\thanks{\copyright~2025 IEEE. Personal use of this material is permitted. Permission from
IEEE must be obtained for all other uses, in any current or future media, including
reprinting/republishing this material for advertising or promotional purposes, creating new
collective works, for resale or redistribution to servers or lists, or reuse of any
copyrighted component of this work in other works. DOI: 10.1109/ICASSP49660.2025.10890544}
}
\maketitle
\begin{abstract}
The integration of renewable energy sources into the electrical grid introduces complex challenges in fault detection and coordination of grid recovery mechanisms. Traditional relay protection systems, which operate based on static rules and predefined thresholds, are inadequate for addressing these challenges, particularly in detecting and isolating faults such as short circuits. Consequently, the conventional methodologies applied to electrical network protection frequently fail to achieve optimal performance in fault detection, especially in terms of adherence to safety standards and the selective limitation of damage. Recent research indicates that machine learning (ML)-based approaches can effectively tackle these issues; however, variations in grid configurations and analysis windows have impeded consistent comparative assessments. In this study, we assess the efficacy of various ML models in detecting electrical faults and pinpointing defective transmission lines within a 10\,ms measurement interval—a critical time-frame for real-time operational viability, for the first time. The most effective model attained an F1 score of 0.991$\pm$0.018 and demonstrated a processing time of 0.342ms$\pm$0.509ms.
\end{abstract}
\section{Introduction}
\label{sec:intro}

In the context of the global drive for decarbonization, the accelerated integration of renewable and distributed energy sources is leading to significant changes in the electrical grid, with the potential to introduce unprecedented complexity and heighten the risk of faults~\cite{papadis_challenges_2020, vde_verband_der_elektrotechnik_elektronik_informationstechnik_ev_zellulare_2015, protection_and_automation_b5_protection_2015}. The transmission line, a key component of the grid, is vulnerable to faults such as short circuits, equipment malfunctions, operator errors, and overloads. Power system protection plays a critical role in preventing outages by detecting these short circuits~\cite{chen_electrical_2005}.

However, the increasing integration of renewable energy sources into existing electrical grids has introduced new challenges such as high degrees of meshing~\cite{biller_protection_2022} or hybrid arrangements of AC and DC lines~\cite{prommetta_protection_2020}.
As a result, it has become increasingly difficult to ensure optimal performance in the areas of fault detection, sensitivity, and selectivity~\cite{vaish_machine_2021} by using traditional protection methods.
These methods rely on fixed rules and thresholds to detect high currents caused by short circuits but struggle with nonlinear classification~\cite{protection_and_automation_b5_protection_2015}.
However, fault currents from renewable energy sources differ from those in traditional energy sources, affecting short-circuit levels and characteristics and therefore potentially causing malfunctions in conventional protection~\cite{chen_electrical_2005}.
Conventional protection systems are also limited by their reliance on predefined scenarios, reducing their ability to handle unfamiliar conditions~\cite{protection_and_automation_b5_protection_2015}.
Historically, protection systems have been self-contained, with little reliance on external measurement ~\cite{adamiak_wide_2006}.
New systems have emerged as a solution based on the IEC 61850 communication standard~\cite{international_electrotechnical_commission_iec_2024}, which leverage remote measurements from all protection systems to enhance grid protection, particularly in transmission lines.
Deploying machine learning (ML) techniques in the electrical grid protection domain confers advantages, including the capacity for non-linear classification, the acquisition of general competencies, and the ability to classify new network scenarios.

Fault detection, fault line identification, and fault localization are distinct yet interconnected tasks in power system protection, crucial for maintaining the reliability and safety of electrical systems~\cite{ieee_power_and_energy_society_ieee_2015}. Fault detection involves recognizing anomalous conditions, such as short circuits, that could disrupt power supply. Following this, fault line identification determines the specific section of the electrical grid where the fault occurs. In contrast, fault localization pinpoints the exact location of the fault along a line segment. This study focuses exclusively on fault detection and fault line identification, excluding fault localization.

The speed of fault detection is a critical factor in ensuring the prompt clearing of faults, which is essential for maintaining reliable and safe power system operation~\cite{gonen_electric_2015}.
The VDEW (Verband der Elektrizitätswirtschaft) sets specific response time requirements for protective systems in case of critical three-phase short-circuits: for extra-high voltage networks, the detection time should not exceed 25\,ms, for high voltage networks 30\,ms, and for medium voltage networks 40\,ms~\cite{ziegler_digitaler_2008}.
The IEEE guide for determining fault location states that relays usually need to detect faults within 10\,ms to 50\,ms~\cite{ieee_power_and_energy_society_ieee_2015}.

A substantial body of research focuses on fault detection in electrical grids. Sapountzoglou et al.~\cite{sapountzoglou_generalizable_2020} presented a deep learning (DL) method to detect and locate faults in low-voltage distribution grids, analyzing the grid's state \(\qty{150}{\milli\second}\) post-fault. Najafzadeh et al.~\cite{najafzadeh_fault_2024} proposed a fault detection approach using a fuzzy logic model, applying thresholds derived from frequency signals collected by phase measurement units (PMUs). 

Hou et al.~\cite{hou_deep-learning-based_2022} introduced a fault type classification method utilizing image augmentation. Rizeakos et al.~\cite{rizeakos_deep_2023} demonstrated a DL application for fault location identification and type classification in active distribution grids, processing data in 20-second batches. Mbey et al.~\cite{mbey_fault_2023} presented DL-based fault detection and classification methods combined with a fuzzy algorithm for smart distribution grids, incorporating virtual smart meters for enhanced data collection. Lastly, Kumar and Kundu~\cite{kumar_faulted_2023} proposed an unsupervised ML approach for fault detection, employing K-Means clustering to differentiate between faulted and non-faulted lines.

Despite significant advancements in ML for detecting and accurately identifying faulty transmission lines in electrical grids, current approaches remain difficult to compare due to inherent differences in data simulation, preprocessing methods, and target metrics. Additionally, many studies do not accurately reflect realistic grid conditions, particularly regarding the real-world boundary conditions in grid topology. Last but not least, none of the published ML-based methods has been tested with the lower bound of the real-time condition, i.e., 10ms, has not been evaluated to the best of our knowledge. 

This paper presents a systematic evaluation of ML models for fault detection and line identification in electrical power grids, focusing on three-phase short circuits in transmission lines. To ensure ML models can generalize well from a synthetic to a later real world dataset, domain randomization is applied to the generation of the electrical grid parameters. As speed is crucial for this task in protection relays, we compare each model's fault detection performance over context windows ranging from 10\,ms to 50\,ms and analyze their runtime.

\section{Methodology}
\label{sec:methodology}

\subsection{Grid Topology and Data Generation} 

The dataset utilized in this study is generated using \mbox{DIgSILENT's} PowerFactory software\footnote{\url{https://www.digsilent.de/en/}}, with the simulation conducted by an expert in electrical power engineering using an extended version of the presented framework in~\cite{wang_generic_2022}. PowerFactory uses physics-based simulations to model complex interactions in electrical power systems, including transmission lines, loads, and fault conditions. To generate realistic data, we simulate grid dynamics with electromagnetic transient simulations. This provides instantaneous voltage and current values, which directly feed into our models, eliminating the need for transformation into the phasor-based values typically used in conventional protection systems.

\begin{figure}[tbp]
\centerline{\includegraphics[width=\linewidth]{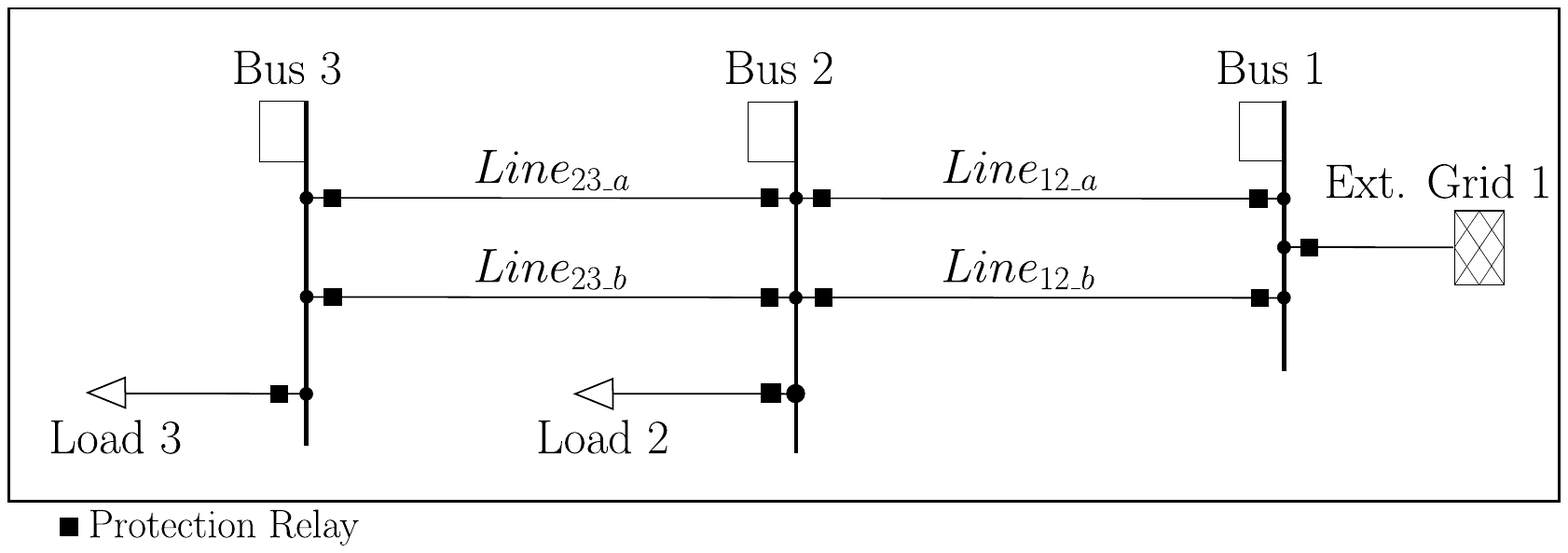}}
\caption{``{Double Line}'' electrical grid topology that is used for data simulation generated based on the network parameters in Tab.~\ref{tab:param_var_overview}}
\label{Netzmodell}
\end{figure}

The electrical grid simulation model is based on the ``\mbox{Double Line}'' topology illustrated in Fig.~\ref{Netzmodell}, a typical topology used for grid protection tests like in~\cite{meyer_hybrid_2020}.
Key parameters of interest, such as settings from the external grid, loads, transmission line lengths, fault initiation time, fault duration, and the location of short circuits on transmission lines,were varied systematically.
This study focuses exclusively on three-phase faults which are the most severe but also less frequent, representing only 5\% of the fault occurrences~\cite{gonen_electric_2015}.

Each simulation, designated as an episode, is characterized by a distinct network and fault parameter configuration. The duration of each episode is \( 1 \, \si{\second} \). The sampling interval is set to \( 50 \, \si{\micro\second} \), which yields 20,000 time steps per episode. The fault initiation can occur at any time between \( 0.2 \, \si{\second} \) and \( 0.5 \, \si{\second} \), with fault durations ranging from \( 5 \, \si{\milli\second} \) to \( 600 \, \si{\milli\second} \) and fault resistance varying from  \( 0.1 \, \si{\ohm} \) to \( 10 \, \si{\ohm} \). Additionally, both $b$ lines are occasionally deactivated to ensure model robustness against topology changes.

In order to guarantee the reliability of each model in response to a variety of scenarios, the electrical grid simulation parameters presented in Tab.~\ref{tab:param_var_overview} are randomized using uniform distributions in accordance with the framework outlined in~\cite{wang_generic_2022}.
The rationale behind randomizing grid parameters, i.e., domain randomization, is to enhance the robustness of ML training and facilitate the domain shift from simulation to reality.
The goal of the chosen parameters is to simulate adequately realistic grid, while covering a wide range of parameters, we based the range of the parameters on~\cite{roeper_kurzschlusstrome_1984, oeding_elektrische_2016}.
The short circuit power was chosen to be in a relatively low range, representing future grids with inverter based resources.
The angle $\phi$ of the external grid is randomized to avoid an alignment of the sinus waves across multiple simulations.
To ensure the realism of the generated models, we ensure that the \(R/X\)-ratio of all lines falls between \(0.05\) and \(0.5\), the load flow converges, and the simulation is numerically stable before, during and after the short circuit.
The variability within these parameter ranges represents the operational diversity inherent in real world electrical grids.

\begin{table}[htbp]
\centering
\caption{Overview of parameter variation of the grid model.}
\resizebox{\linewidth}{!}{
\begin{tabular}{|c|c|c|c|}
\hline
\textbf{Element} & \textbf{Parameter} & \textbf{Min. Value} &  \textbf{Max. Value}\\
\hline
Line  & Length (km)  & 10  & 60  \\
Line  & Reactance $X'$ ($\Omega/km$)  & 0.35  & 0.45 \\
Line  & Resistance $R'$ ($\Omega/km$)  & 0.01  & 0.20 \\
Line  & Capacitance $C'$ ($nF/km$) & 8.50  & 10 \\
Load  & $P$ (MW) & 20  & 50 \\
Load  & $Q$ (MVar) & -20  & 20 \\
Ext. Grid & Shc. Power $S_k''$ (MVA) & 90  & 1000 \\
Ext. Grid & Voltage Setpoint $V_{set}$ (pu.) & 0.95  & 1.05 \\
Ext. Grid & Angle $\phi$ (deg) & -180  & 180 \\
\hline
\end{tabular}}
\label{tab:param_var_overview}
\end{table}

The ``Double-Line'' network model visualized in Fig.~\ref{Netzmodell} consists of one external grid, two loads, and three buses connected by four transmission lines.
Each transmission line is equipped with two protection relay (PR) devices, one at each end, measuring instantaneous values of current and voltage across three phases \((A, B, C)\).
The current and voltage measurements from the PR devices are represented as:
\begin{equation}
    I_{PR}(t) = (I_A(t), I_B(t), I_C(t)), \quad t \in [0, 1]\, \text{s}
\end{equation}
\begin{equation}
    V_{PR}(t) = (V_A(t), V_B(t), V_C(t)), \quad t \in [0, 1]\, \text{s}
\end{equation}

Here, \(I\) represents current and \(V\) represents voltage, with subscripts \(A\), \(B\), and \(C\) corresponding to the three phases.
So on each transmission line two PR devices record voltage and current per time step, forming a multivariate time series represented as:

\begin{equation}
    X_{Line}(t) = \begin{bmatrix} I_{PR\_1}(t) \\ V_{PR\_1}(t) \\ I_{PR\_2}(t)  \\ V_{PR\_1}(t) \end{bmatrix}, \quad t \in [0, 1]\, \text{s}
\label{eq:X_line}
\end{equation}

 With two relay devices per transmission line, this results in twelve measurements per line. For four transmission lines, a total of 48 measurements are collected per time step, comprising a multivariate time series across all PR devices.

For each episode, we collect the labels and the corresponding electrical grid parameters. The labels consist of two components: \( \textit{fault\_start} \), which represents the time of the fault event (in seconds), and \( \textit{fault\_line} \), a categorical variable indicating the affected transmission line. In this context, \(\textit{fault\_start} \) is used for fault detection, while \( \textit{fault\_line} \) is leveraged for fault line identification.

\subsection{Data Preprocessing} 

To effectively train our models for fault detection, we preprocess the raw simulation data by trimming each episode to a range of \(\pm 80 \, \text{ms}\) around the fault start. This range is chosen to capture critical events just before and after the fault occurs. 
To simulate real-time constraints in a protection relay, we slide across the time series with a step size of 5\,ms, ensuring that each snippet overlaps at least two segments.
We evaluate the impact of varying window lengths: 10\,ms, 20\,ms, 30\,ms, 40\,ms, and 50\,ms.
A detailed overview of the number of time steps corresponding to these lengths is provided in Tab.~\ref{tab:window_length_overview}.

\begin{table}[htbp]
\centering
\caption{Overview of window lengths and resulting timesteps per window, number of windows, number of windows containing a fault, and number of features per window.}
\resizebox{0.98\linewidth}{!}{
\begin{tabular}{|c|c|c|c|c|}
\hline
\textbf{Window} & \textbf{Timesteps} & \textbf{\# Windows} &  \textbf{\# Fault} & \textbf{\# Features} \\
\textbf{Length (ms)} & \textbf{/ Window} & & \textbf{Windows} & \textbf{/ Window} \\
\hline
10  & 200  & 15500  & 500  & 9600 \\
20  & 400  & 14500  & 1500 & 19200 \\
30  & 600  & 13500  & 2500 & 28800 \\
40  & 800  & 12500  & 3500 & 38400 \\
50  & 1000 & 11500  & 4500 & 48000 \\
\hline
\end{tabular}}

\label{tab:window_length_overview}
\end{table}

\subsection{Machine Learning Models for Fault Detection and Line Identification}

In this work, fault detection is treated as a binary classification task, while line identification is handled as a multi-class classification problem with four distinct categories. The feature space consists of a vector formed by concatenating simulated voltage and current data from the four transmission lines, as introduced in Eq.~\ref{eq:X_line}, over a specified window length. The number of features depends on the window length (see Tab.~\ref{tab:window_length_overview}). The input is a multivariate time series representing voltage and current data that captures grid behavior over this period.

To formulate the fault detection task as a binary classification problem, we label each window using \( \textit{fault\_start} \) to indicate whether a three-phase short circuit fault occurs within the window. Specifically, a fault label is assigned if the condition \( t_{\text{start}} + \epsilon < \textit{fault\_start} < t_{\text{end}} - \epsilon \) is met, where \( t_{\text{start}} \) and \( t_{\text{end}} \) represent the window's start and end timestamps, and \( \epsilon = 5 \, \mu\text{s} \) ensures that the fault event is entirely contained within the window. If this condition is true, the label is 1; otherwise, it is 0.

Additionally, each window is associated with a fault line, represented as a categorical variable that indicates the affected transmission line where the fault occurred. The fault line is assigned as expressed in Eq.~\ref{eq:fault_line}.

\begin{equation}
    \textit{fault\_line} \in 
\left\{ 
\text{Line}_{12\_a}, \, 
\text{Line}_{12\_b}, \, 
\text{Line}_{23\_a}, \, 
\text{Line}_{23\_b} 
\right\}
\label{eq:fault_line}
\end{equation}

To ensure a comprehensive evaluation, a diverse set of classifiers that are proposed in the literature such as Logistic Regression (LG), Ridge, Stochastic Gradient Descent (SGD), K-Nearest Neighbors (KNN) employed with \(K=5\), Multi-Layer Perceptron (MLP), and Support Vector Machines (SVM), alongside ensemble methods like AdaBoost, Bagging, ExtraTrees (ET), Histogram-based Gradient Boosting (GB), Random Forest (RF), Stacking, and Voting Classifiers are evaluated. All models are implemented using \textit{scikit-learn}~\cite{scikit_learn_2015}.

A 10-fold cross-validation is employed in all experiments, where each iteration utilizes a 9:1 training-to-test split to assess the robustness of the models. This ensures that each model is tested on unseen data during each iteration. Features are standardized by removing the mean and scaling to unit variance for uniform input representation. Evaluation metrics such as accuracy, precision, recall, sensitivity, specificity, and F1 score are calculated, with the F1 score reported due to its balance between sensitivity and specificity.

For fault detection, all extracted windows were used (Tab.~\ref{tab:window_length_overview}), but for fault line identification, only windows that contain faults were employed, as line identification is only possible after detecting a fault.
To compare model run-time, the time taken for scaling and predicting a single window was measured over 5000 iterations, with mean and standard deviation reported.

\section{Experiments and Results}
\label{sec:results}
\begin{figure}[htbp]
\centerline{\includegraphics[width=0.98\linewidth]{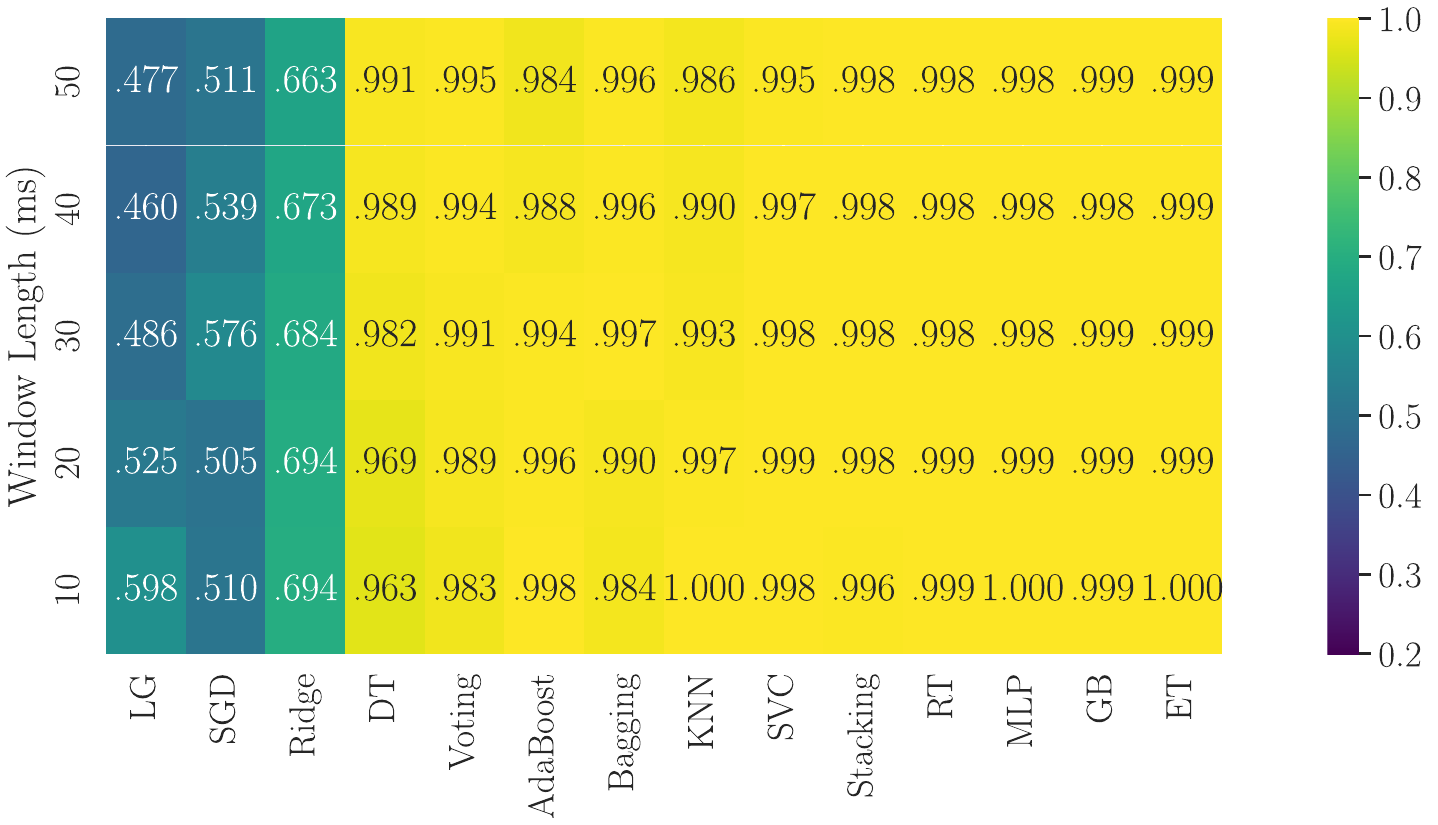}}
\caption{Heatmap of the Fault Detection F1 Scores}
\label{heatmap_fault}
\end{figure}

The results of the fault detection evaluation are shown in Fig.~\ref{heatmap_fault}. Of the 14 models tested, 11 achieved F1 scores above 0.96, regardless of the window size. However, LG, Ridge, and SDG performed poorly, with F1 scores below 0.69. The best-performing models--ET, GB, MLP, RF, Stacking, and SVM--each achieved F1 scores up to 0.99 over all window lengths.

\begin{figure}[htbp]
\centerline{\includegraphics[width=0.98\linewidth]{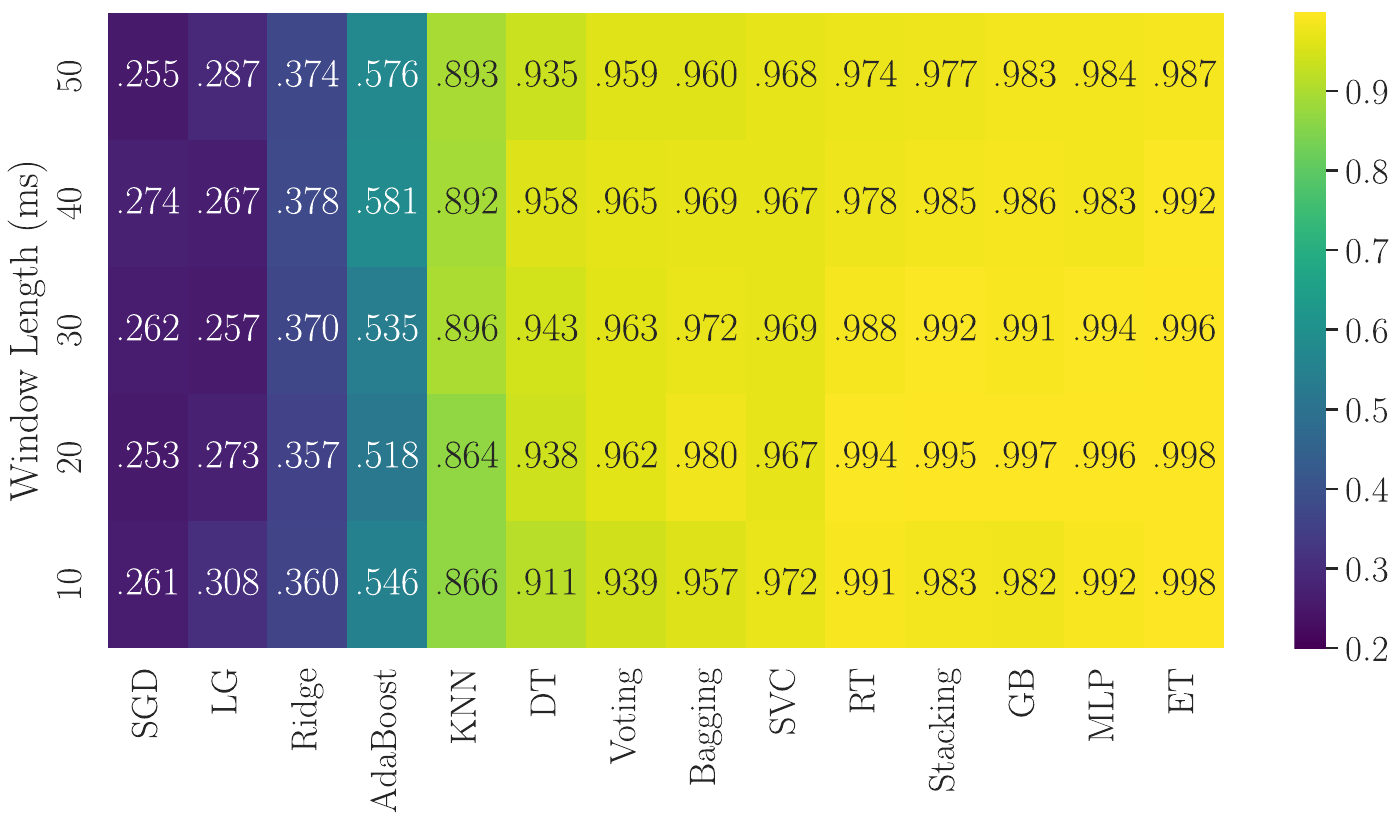}}
\caption{Heatmap of the Fault Line Identification F1 Scores}
\label{heatmap_fault_target}
\end{figure}

Fig.~\ref{heatmap_fault_target} presents the results of the fault line identification task. Out of the 14 models tested, six achieved a mean F1 score above 0.96. In contrast, LG, Ridge, SDG, and AdaBoost underperformed (F1 scores$\leq$0.58). More models struggled in this task compared to fault detection. The top-performing models--ET, GB, MLP, RF, and Stacking--consistently achieved a mean F1 score of up to 0.97.

\begin{figure}[htbp]
\centerline{\includegraphics[width=\linewidth]{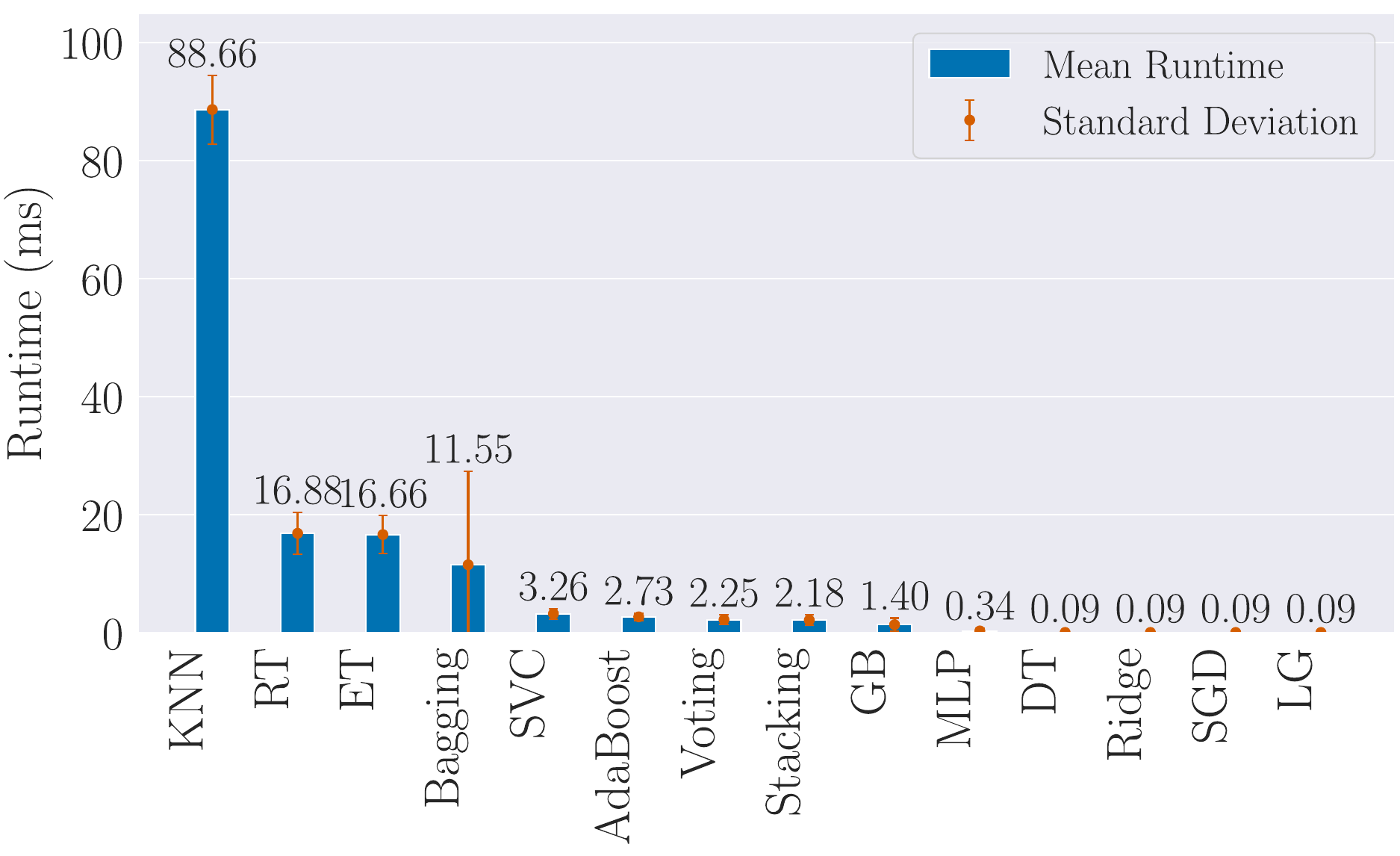}}
\caption{Overview of Mean and Std. Runtime of each ML Model}
\label{overview_runtime_ml_models}
\end{figure}

The runtime experiments are carried out on an Intel® Core™ i7-13700K processor using standard Python. Each model runtime is tested by running 5000 iterations. The results are presented in Fig.~\ref{overview_runtime_ml_models}. The three fastest models were Ridge, SDG and LG. Additionally, the KNN (\(K=5\)) had the slowest runtime of 88.6\(\pm\)5.8\,ms a stark outlier compared to the rest.

\section{Discussion}\label{sec:discussion}

The results indicate that most models are effective regardless of the window length.
Given the shortest window length of 10\,ms the top-performing models -- ET, GB, MLP, RF, Stacking, and SVM -- achieved mean F1 scores of up to 0.99 for the fault detection and 0.98 for the fault line identification. The five best models can reliably detect faults and identify faulty lines using a measurement window equal to half a period at the standard frequency of \(\qty{50}{\hertz}\).

The three fastest models performed the worst in both tasks, indicating that these models are too simple to detect or identify the fault. The fourth-fastest model was the Decision Tree with a runtime of \(\qty{0.09}{\milli\second}\), its mean F1 scores of 0.96 and 0.91 for the fault detection and line identification tasks, respectively, were relatively low.
The results indicate that the MLP, GB, and Stacking are the most effective models, exhibiting the highest F1 scores for both fault detection and line identification while also demonstrating competitive runtimes. The runtime for the MLP was 0.34\,ms, the GB had a runtime of 1.40\,ms and for the Stacking was 2.18\,ms.

\section{Conclusion}
\label{sec:conclusion}
This study presents a systematic evaluation of ML models for the detection of faults and identification of lines in electrical power grids, with a particular focus on three-phase short circuits in transmission lines.
The results demonstrate that, in the specified scenarios, the majority of the evaluated models achieve exceptional scores when a window length of 10 ms is employed. Although discrepancies in recorded runtime were observed, this is dependent on the hardware utilized and warrants further investigation.
The study is limited to a single grid, but future research will explore the transferability of pre-trained models to different grids and extend the evaluation to other short-circuit types, including two-phase and grounding faults. Future work will also examine various grid topologies, load conditions, energy sources, multiple fault scenarios, and the models' resilience to noisy or incomplete data to ensure reliability in diverse conditions. 

\section*{Acknowledgment}
This project was funded by the Deutsche Forschungsgemeinschaft (DFG, German Research Foundation) - 535389056.

\bibliographystyle{IEEEtran}
\bibliography{refs}

@article{mbey_fault_2023,
	title = {Fault detection and classification using deep learning method and neuro‐fuzzy algorithm in a smart distribution grid},
	volume = {2023},
	issn = {2051-3305, 2051-3305},
	url = {https://ietresearch.onlinelibrary.wiley.com/doi/10.1049/tje2.12324}    ,
	doi = {10.1049/tje2.12324},
	pages = {e12324},
	number = {11},
	journal = {The Journal of Engineering},
	shortjournal = {The Journal of Engineering},
	author = {Mbey, Camille Franklin and Foba Kakeu, Vinny Junior and Boum, Alexandre Teplaira and Souhe, Felix Ghislain Yem},
	urldate = {2024-09-04},
	date = {2023-11},
	langid = {english},
    year = {2023},
}

@article{rizeakos_deep_2023,
	title = {Deep learning-based application for fault location identification and type classification in active distribution grids},
	volume = {338},
	issn = {03062619},
	url = {https://linkinghub.elsevier.com/retrieve/pii/S0306261923002969},
	doi = {10.1016/j.apenergy.2023.120932},
	pages = {120932},
	journal = {Applied Energy},
	shortjournal = {Applied Energy},
	author = {Rizeakos, V. and Bachoumis, A. and Andriopoulos, N. and Birbas, M. and Birbas, A.},
	urldate = {2024-09-04},
	date = {2023-05},
	langid = {english},
    year = {2023},
}

@article{hou_deep-learning-based_2022,
	title = {Deep-Learning-Based Fault Type Identification Using Modified {CEEMDAN} and Image Augmentation in Distribution Power Grid},
	volume = {22},
	rights = {https://ieeexplore.ieee.org/Xplorehelp/downloads/license-information/{IEEE}.html},
	issn = {1530-437X, 1558-1748, 2379-9153},
	url = {https://ieeexplore.ieee.org/document/9638639/},
	doi = {10.1109/JSEN.2021.3133352},
	pages = {1583--1596},
	number = {2},
	journal = {{IEEE} Sensors Journal},
	shortjournal = {{IEEE} Sensors J.},
	author = {Hou, Si-Zu and Guo, Wei and Wang, Zi-Qi and Liu, Ya-Ting},
	urldate = {2024-09-04},
	date = {2022-01-15},
    year = {2022},
}

@book{gonen_electric_2015,
	edition = {3rd edition.},
	title = {Electric Power Distribution Engineering, 3rd Edition},
	pagetotal = {1061},
	publisher = {{CRC} Press},
	author = {Gonen, Turan and {Safari, an O'Reilly Media Company.}},
	date = {2015},
	note = {{OCLC}: 1105773230},
    year = {2015},
}

@article{najafzadeh_fault_2024,
	title = {Fault Detection, Classification and Localization Along the Power Grid Line Using Optimized Machine Learning Algorithms},
	volume = {17},
	issn = {1875-6883},
	url = {https://link.springer.com/10.1007/s44196-024-00434-7}    ,
	doi = {10.1007/s44196-024-00434-7},
	pages = {49},
	number = {1},
	journal = {International Journal of Computational Intelligence Systems},
	shortjournal = {Int J Comput Intell Syst},
	author = {Najafzadeh, Masoud and Pouladi, Jaber and Daghigh, Ali and Beiza, Jamal and Abedinzade, Taher},
	urldate = {2024-09-03},
	date = {2024-03-25},
	langid = {english},
    year = {2024},
}

@article{sapountzoglou_generalizable_2020,
	title = {A generalizable and sensor-independent deep learning method for fault detection and location in low-voltage distribution grids},
	volume = {276},
	issn = {03062619},
	url = {https://linkinghub.elsevier.com/retrieve/pii/S0306261920308114},
	doi = {10.1016/j.apenergy.2020.115299},
	pages = {115299},
	journal = {Applied Energy},
	shortjournal = {Applied Energy},
	author = {Sapountzoglou, Nikolaos and Lago, Jesus and De Schutter, Bart and Raison, Bertrand},
	urldate = {2024-09-03},
	date = {2020-10},
	langid = {english},
    year = {2020},
}

@misc{protection_and_automation_b5_protection_2015,
	title = {Protection of distribution systems with distributed energy resources},
	url = {https://www.e-cigre.org/publications/detail/613-protection-of-distribution-systems-with-distributed-energy-resources.html},
	type = {Technical Brochure},
	author = {{Protection and automation (B5)} and Active distribution systems {and} distributed energy resources (C6)},
	urldate = {2024-09-03},
	date = {2015},
	langid = {british},
    year = {2015},
}

@article{vaish_machine_2021,
	title = {Machine learning applications in power system fault diagnosis: Research advancements and perspectives},
	volume = {106},
	issn = {09521976},
	url = {https://linkinghub.elsevier.com/retrieve/pii/S0952197621003523},
	doi = {10.1016/j.engappai.2021.104504},
	shorttitle = {Machine learning applications in power system fault diagnosis},
	pages = {104504},
	journal = {Engineering Applications of Artificial Intelligence},
	shortjournal = {Engineering Applications of Artificial Intelligence},
	author = {Vaish, Rachna and Dwivedi, U.D. and Tewari, Saurabh and Tripathi, S.M.},
	urldate = {2024-09-03},
	date = {2021-11},
	langid = {english},
    year = {2021},
}

@article{adamiak_wide_2006,
	title = {Wide Area Protection—Technology and Infrastructures},
	volume = {21},
	rights = {https://ieeexplore.ieee.org/Xplorehelp/downloads/license-information/{IEEE}.html},
	issn = {0885-8977},
	url = {http://ieeexplore.ieee.org/document/1610668/},
	doi = {10.1109/TPWRD.2005.855481},
	pages = {601--609},
	number = {2},
	journal = {{IEEE} Transactions on Power Delivery},
	shortjournal = {{IEEE} Trans. Power Delivery},
	author = {Adamiak, M.G. and Apostolov, A.P. and Begovic, M.M. and Henville, C.F. and Martin, K.E. and Michel, G.L. and Phadke, A.G. and Thorp, J.S.},
	urldate = {2024-09-03},
	date = {2006-04},
	langid = {english},
    year = {2006},
}

@book{chen_electrical_2005,
	location = {Boston},
	title = {The electrical engineering handbook},
	isbn = {9781417552665},
	pagetotal = {1208},
	publisher = {Elsevier Academic Press},
	author = {Chen, Wai-Kai},
	date = {2005},
	note = {{OCLC}: 57371415},
    year = {2005},
}

@misc{international_electrotechnical_commission_iec_2024,
	title = {{IEC} 61850:2024 {SER} {\textbar} {IEC}},
	url = {https://webstore.iec.ch/en/publication/6028},
	version = {1.0},
	author = {International Electrotechnical Commission},
	urldate = {2024-09-03},
	date = {2024-05-02},
    year={2024},
}

@misc{vde_verband_der_elektrotechnik_elektronik_informationstechnik_ev_zellulare_2015,
	title = {Der Zellulare Ansatz - {VDE} Studie},
	url = {www.vde.com/studie-zellularer-ansatz},
	institution = {{VDE} Verband der Elektrotechnik Elektronik Informationstechnik e.V.},
	type = {Fachinformation},
	author = {{VDE Verband der Elektrotechnik Elektronik Informationstechnik e.V.}},
	date = {2015-06-01},
    year={2015},
}

@article{papadis_challenges_2020,
	title = {Challenges in the decarbonization of the energy sector},
	volume = {205},
	issn = {03605442},
	url = {https://linkinghub.elsevier.com/retrieve/pii/S0360544220311324},
	doi = {10.1016/j.energy.2020.118025},
	pages = {118025},
    journal = {Energy},
	shortjournal = {Energy},
	author = {Papadis, Elisa and Tsatsaronis, George},
	urldate = {2024-09-04},
	date = {2020-08},
	langid = {english},
    year={2020},
}

@inproceedings{kumar_faulted_2023,
	location = {Cox's Bazar, Bangladesh},
	title = {Faulted Line Identification in Power Network using Unsupervised Machine Learning},
	rights = {https://doi.org/10.15223/policy-029}             ,
	isbn = {9798350318739},
	url = {https://ieeexplore.ieee.org/document/10428707/},
	doi = {10.1109/ICPS60393.2023.10428707},
	eventtitle = {2023 10th {IEEE} International Conference on Power Systems ({ICPS})},
	pages = {1--6},
	booktitle = {2023 10th {IEEE} International Conference on Power Systems ({ICPS})},
	publisher = {{IEEE}},
	author = {Kumar, A Rupesh and Kundu, Pratim},
	urldate = {2024-09-06},
	date = {2023-12-13},
	year = {2023},
}

@misc{ieee_power_and_energy_society_ieee_2015,
	title = {{IEEE} Guide for Determining Fault Location on {AC} Transmission and Distribution Lines},
	url = {http://ieeexplore.ieee.org/document/7024095/},
	doi = {10.1109/IEEESTD.2015.7024095},
	publisher = {{IEEE}},
	author = {{IEEE Power and Energy Society}},
	urldate = {2024-09-08},
	date = {2015-01-30},
}

@article{scikit_learn_2015,
  title={Scikit-learn: Machine Learning in {P}ython},
  author={Pedregosa, F. and Varoquaux, G. and Gramfort, A. and Michel, V.
          and Thirion, B. and Grisel, O. and Blondel, M. and Prettenhofer, P.
          and Weiss, R. and Dubourg, V. and Vanderplas, J. and Passos, A. and
          Cournapeau, D. and Brucher, M. and Perrot, M. and Duchesnay, E.},
  journal={Journal of Machine Learning Research},
  volume={12},
  pages={2825--2830},
  year={2011}
}

@article{biller_protection_2022,
	title = {Protection Algorithms for Closed-Ring Grids With Distributed Generation},
	volume = {37},
	rights = {https://ieeexplore.ieee.org/Xplorehelp/downloads/license-information/{IEEE}.html},
	issn = {0885-8977, 1937-4208},
	url = {https://ieeexplore.ieee.org/document/9684980/},
	doi = {10.1109/TPWRD.2022.3144004},
	pages = {4042--4052},
	number = {5},
	journal = {{IEEE} Transactions on Power Delivery},
	shortjournal = {{IEEE} Trans. Power Delivery},
	author = {Biller, Martin and Jaeger, Johann},
	urldate = {2024-09-09},
	date = {2022-10},
    year = {2022}
}

@article{prommetta_protection_2020,
	title = {Protection Coordination of {AC}/{DC} Intersystem Faults in Hybrid Transmission Grids},
	volume = {35},
	rights = {https://ieeexplore.ieee.org/Xplorehelp/downloads/license-information/{IEEE}.html},
	issn = {0885-8977, 1937-4208},
	url = {https://ieeexplore.ieee.org/document/9112359/},
	doi = {10.1109/TPWRD.2020.3000731},
	pages = {2896--2904},
	number = {6},
	journal = {{IEEE} Transactions on Power Delivery},
	shortjournal = {{IEEE} Trans. Power Delivery},
	author = {Prommetta, Jonas and Schindler, Jakob and Jaeger, Johann and Keil, Timo and Butterer, Christoph and Ebner, Guenter},
	urldate = {2024-09-09},
	date = {2020-12},
    year = {2020}
}

@article{meyer_hybrid_2020,
	title = {Hybrid fuzzy evaluation algorithm for power system protection security assessment},
	volume = {189},
	issn = {03787796},
	url = {https://linkinghub.elsevier.com/retrieve/pii/S037877962030359X},
	doi = {10.1016/j.epsr.2020.106555 },
	pages = {106555},
	journal = {Electric Power Systems Research},
	shortjournal = {Electric Power Systems Research},
	author = {Meyer, Georg Janick and Lorz, Tobias and Wehner, Rene and Jaeger, Johann and Dauer, Maximilian and Krebs, Rainer},
	urldate = {2024-09-10},
	date = {2020-12},
	langid = {english},
    year = {2020}
}

@inproceedings{wang_generic_2022,
	location = {Kassel, Germany},
	title = {A Generic Data Generation Framework for Short Circuit Detection Training of Neural Networks},
	isbn = {978-3-8007-6013-8 },
	url = {https://ieeexplore.ieee.org/document/10104226},
	eventtitle = {{PESS} + {PELSS} 2022; Power and Energy Student Summit},
	pages = {49--54},
	booktitle = {{PESS} + {PELSS} 2022; Power and Energy Student Summit},
	publisher = {{VDE}},
	author = {Wang, Minxiao and Kordowich, Georg and Jäger, Johann},
	date = {2022-11-02},
    year= {2022},
}

@book{roeper_kurzschlusstrome_1984,
	location = {Berlin, München: Siemens},
	edition = {6},
	title = {Kurzschlußströme in Drehstromnetzen},
	isbn = {9783800913855 },
	pagetotal = {171},
	publisher = {Publicis Corporate Publishing},
	author = {Roeper, Richard and {Mitlehner, Friedrich}},
	date = {1984},
    year = {1984},
}

@book{ziegler_digitaler_2008,
	location = {Erlangen},
	edition = {2. Aufl},
	title = {Digitaler Distanzschutz: Grundlagen und Anwendung},
	isbn = {9783895783203 },
	shorttitle = {Digitaler Distanzschutz},
	pagetotal = {392},
	publisher = {Publicis Corp. Publ},
	author = {Ziegler, Gerhard},
	date = {2008},
    year = {2008},
}

@book{oeding_elektrische_2016,
	location = {Berlin, Heidelberg},
	edition = {8},
	title = {Elektrische Kraftwerke und Netze},
	rights = {http://www.springer.com/tdm},
	isbn = {9783662527023},
	url = {http://link.springer.com/10.1007/978-3-662-52703-0},
	pagetotal = {1107},
	publisher = {Springer Berlin Heidelberg},
	author = {Oeding, Dietrich and Oswald, Bernd R.},
	urldate = {2024-09-10},
	date = {2016},
	langid = {german},
	doi = {10.1007/978-3-662-52703-0},
    year = {2016},
}

\end{document}